\documentclass[11pt]{article}
\usepackage{paclic30}
\usepackage{times}
\usepackage{latexsym}
\usepackage{amsmath}
\usepackage{multirow}
\usepackage{url}
\usepackage{graphicx}
\usepackage{caption}
\usepackage{mathtools}

\title{Testing APSyn against Vector Cosine on\\ Similarity Estimation}

\author{%
	Enrico Santus\textsuperscript{[1]}, Emmanuele Chersoni\textsuperscript{[2]}, Alessandro Lenci\textsuperscript{[3]}, Chu-Ren Huang\textsuperscript{[1]}, Philippe Blache\textsuperscript{[2]}\\
	{[1] The Hong Kong Polytechnic University, Hong Kong}\\
	{[2] Aix-Marseille University}\\
	{[3] University of Pisa}\\
	{\tt\{esantus, emmanuelechersoni\}@gmail.com}\\
	{\tt alessandro.lenci@unipi.it}\\
	{\tt churen.huang@polyu.edu.hk}\\
	{\tt blache@lpl-aix.fr}\\
}

	\date{}
\makeindex 
\begin{document}
\maketitle
\begin{abstract}
In Distributional Semantic Models (DSMs), Vector Cosine is widely used to estimate similarity between word vectors, although this measure was noticed to suffer from several shortcomings. The recent literature has proposed other methods which attempt to mitigate such biases. In this paper, we intend to investigate APSyn, a measure that computes the extent of the intersection between the most associated contexts of two target words, weighting it by context relevance. We evaluated this metric in a similarity estimation task on several popular test sets, and our results show that APSyn is in fact highly competitive, even with respect to the results reported in the literature for word embeddings. On top of it, APSyn addresses some of the weaknesses of Vector Cosine, performing well also on genuine similarity estimation.
\end{abstract}

\section{Introduction}

Word similarity is one of the most important and most studied problems in Natural Language Processing (NLP), as it is fundamental for a wide range of tasks, such as \textit{Word Sense Disambiguation} (WSD), \textit{Information Extraction} (IE), \textit{Paraphrase Generation} (PG), as well as the automatic creation of semantic resources.
Most of the current approaches to word similarity estimation rely on some version of the Distributional Hypothesis (DH), which claims that words occurring in the same contexts tend to have similar meanings \cite{harris1954distributional,firth1957papers,sahlgren2008distributional}. Such hypothesis provides the theoretical ground for Distributional Semantic Models (DSMs), which represent word meaning by means of high-dimensional vectors encoding corpus-extracted co-occurrences between targets and their linguistic contexts \cite{turney2010frequency}.\\
Traditional DSMs initialize vectors with co-occurrence frequencies. Statistical measures, such as Positive Pointwise Mutual Information (PPMI) or its variants \cite{church1990word,bullinaria2012extracting,levy2015improving}, have been adopted to normalize these values. Also, these models have exploited the power of dimensionality reduction techniques, such as Singular Value Decomposition (SVD; Landauer and Dumais, 1997) and Random Indexing \cite{sahlgren2005introduction}. \\
These first-generation models are currently referred to as count-based, as distinguished from the context-predicting ones, which have been recently proposed in the literature \cite{bengio2006neural,collobert2008unified,turian2010word,huang2012improving,mikolov2013efficient}. More commonly known as \textit{word embeddings}, these second-generation models learn meaning representations through neural network training: the vectors dimensions are set to maximize the probability for the contexts that typically occur with the target word.\\
Vector Cosine is generally adopted by both types of models as a similarity measure. However, this metric has been found to suffer from several problems \cite{li2013distance,faruqui2016problems}, such as a bias towards features with higher values and the inability of considering how many features are actually shared by the vectors. Finally, Cosine is affected by the hubness effect \cite{dinu2014improving,schnabel2015evaluation}, i.e. the fact that words with high frequency tend to be universal neighbours. Even though other measures have been proposed in the literature \cite{deza2009encyclopedia}, Vector Cosine is still by far the most popular one \cite{turney2010frequency}. However, in a recent paper of Santus et al. \shortcite{santus2016what}, the authors have claimed that Vector Cosine is outperformed by APSyn (Average Precision for Synonymy), a metric based on the extent of the intersection between the most salient contexts of two target words. The measure, tested on a window-based DSM, outperformed Vector Cosine on the ESL and on the TOEFL datasets.\\
In the present work, we perform a systematic evaluation of APSyn, testing it on the most popular test sets for similarity estimation - namely WordSim-353 \cite{finkelstein2001placing}, MEN \cite{bruni2014multimodal} and SimLex-999 \cite{hill2015simlex}. For comparison, Vector Cosine is also calculated on several count-based DSMs. We implement a total of twenty-eight models with different parameters settings, each of which differs according to corpus size, context window width, weighting scheme and SVD application. The new metric is shown to outperform Vector Cosine in most settings, except when the latter metric is applied on a PPMI-SVD reduced matrix \cite{bullinaria2012extracting}, against which APSyn still obtains competitive performances. The results are also discussed in relation to the state-of-the-art DSMs, as reported in Hill et al. \shortcite{hill2015simlex}. In such comparison, the best settings of our models outperform the word embeddings in almost all datasets. A pilot study was also carried out to investigate whether APSyn is scalable. Results prove its high performance also when calculated on large corpora, such as those used by Baroni et al. \shortcite{baroni2014don}.\\
On top of the performance, APSyn seems not to be subject to some of the biases that affect Vector Cosine.
Finally, considering the debate about the ability of DSMs to calculate genuine similarity as opposed to word relatedness \cite{turney2001mining,agirre2009study,hill2015simlex}, we test the ability of the models to quantify genuine semantic similarity.

\section{Background}

\subsection{DSMs, Measures of Association and Dimensionality Reduction}

Count-based DSMs are built in an unsupervised way. Starting from large preprocessed corpora, a matrix \textit{$M_{(m \times n)}$} is built, in which each row is a vector representing a target word in a vocabulary of size \textit{m}, and each column is one of the \textit{n} potential contexts \cite{turney2010frequency,levy2015improving}. The vector dimensions are counters recording how many times the contexts co-occur with the target words.\\
Since raw frequency is highly skewed, most DSMs have adopted more sophisticated association measures, such as Positive PMI (PPMI; Church and Hanks, 1990; Bullinaria and Levy, 2012; Levy et al., 2015) and Local Mutual Information (LMI; Evert, 2005). PPMI compares the observed joint probability of co-occurrence of \textit{w} and \textit{c} with their probability of co-occurrence assuming statistical indipendence. It is defined as:
 
\begin{equation} \label{1}
PPMI(w,c) = max(PMI(w,c), 0)
\end{equation}

\begin{equation} \label{2}
PMI(w,c) = log \left (\frac{P(w,c)}{P(w)Ã—P(c)}\right ) = log\left ( \frac{|w,c|Ã—D}{|w|Ã—|c|}\right )
\end{equation}

where \textit{w} is the target word, \textit{c} is the given context, \textit{P(w,c)} is the probability of co-occurrence, and \textit{D} is the collection of observed word-context pairs.\\
Unlike frequency, PPMI was found to have a bias towards rare events. LMI could therefore be used to reduce such bias and it consists in multiplying the PPMI of the pair by its co-occurrence frequency.
Since target words may occur in hundreds of thousands contexts, most of which are not informative, methods for dimensionality reduction have been investigated, such as truncated SVD \cite{deerwester1990indexing,landauer1997solution,turney2010frequency,levy2015improving}. SVD has been regarded as a method for noise reduction and for the discovery of latent dimensions of meaning, and it has been shown to improve similarity measurements when combined with PPMI \cite{bullinaria2012extracting,levy2015improving}.
As we will see in the next section, APSyn applies another type of reduction, which consists in selecting only the top-ranked contexts in a relevance sorted context list for each word vector. Such reduction complies with the principle of cognitive economy (i.e. only the most relevant contexts are elaborated; see Finton, 2002) and with the results of behavioural studies, which supported feature saliency \cite{smith1974structure}. Since APSyn was defined for linguistic contexts \cite{santus2016what}, we did not test it on SVD-reduced spaces, leaving such test to further studies.

\subsection{Similarity Measures}
Vector Cosine, by far the most common distributional similarity metric \cite{turney2010frequency,landauer1997solution,jarmasz2004roget,mikolov2013efficient,levy2015improving}, looks at the normalized correlation between the dimensions of two word vectors, \textit{$w_1$} and \textit{$w_2$} and returns a score between -1 and 1. It is described by the following equation:
 
\begin{equation} \label{4}
cos({w_1},{w_2}) = \frac{\sum_{i=1}^{n}f_1i \times f_{2i}}{\sqrt{\sum_{i=1}^{n}f_1i} \times \sqrt{\sum_{i=1}^{n}f_2i}}
\end{equation} 
where \textit{$f_ix$} is the \textit{i}-th dimension in the vector \textit{x}.\\
Despite its extensive usage, Vector Cosine has been recently criticized for its hyper sensibility to features with high values and for the inability of identifying the actual feature intersection \cite{li2013distance,schnabel2015evaluation}. Recalling an example by Li and Han \shortcite{li2013distance}, the Vector Cosine for the toy-vectors \textit{$a=[1,2,0]$} and \textit{$b=[0,1,0]$} (i.e. 0.8944) is unexpectedly higher than the one for \textit{a} and \textit{$c=[2,1,0]$} (i.e. 0.8000), and even higher than the one for the toy-vectors \textit{a} and \textit{$d=[1,2,1]$} (i.e. 0.6325), which instead share a larger feature intersection. Since the Vector Cosine is a distance measure, it is also subject to the hubness problem, which was shown by Radovanovic et al. \shortcite{radovanovic2010existence} to be an inherent property of data distributions in high-dimensional vector space. The problem consists in the fact that vectors with high frequency tend to get high scores with a large number of other vectors, thus becoming universal nearest neighbours \cite{dinu2014improving,schnabel2015evaluation,faruqui2016problems}. \\
Another measure of word similarity named APSyn \footnote{Scripts and information can be found at https://github.com/esantus/APSyn} has been recently introduced in Santus et al. \shortcite{santus2016unsupervised} and Santus et al. \shortcite{santus2016what}, and it was shown to outperform the vector cosine on the TOEFL \cite{landauer1997solution} and on the ESL \cite{turney2001mining} test sets. This measure is based on the hypothesis that words carrying similar meanings share their most relevant contexts in higher proportion compared to less similar words. The authors define APSyn as the extent of the weighted intersection between the top most salient contexts of the target words, weighting it by the average rank of the intersected features in the PPMI-sorted contexts lists of the target words:

\begin{equation} \label{5} \small
\begin{multlined}
APSyn({w_1},{w_2}) = \\ 
\\
\sum_{f\epsilon N(F_1)\cap N(F_2)}{\frac{1}{(rank_1(f)+rank_2(f))/2}} 
\end{multlined}
\end{equation} 
meaning: for every feature \textit{f} included in the intersection between the top \textit{N} features of \textit{$w_1$} and the top of \textit{$w_2$} (i.e. \textit{$N(f_1)$} and \textit{$N(f_2)$}), add 1 divided by the average rank of the feature in the PPMI-ranked features of \textit{$w_1$} (i.e. \textit{$rank_1$}) and \textit{$w_2$} (i.e. \textit{$rank_2$}). According to the authors, \textit{N} is a parameter, generally ranging between 100 and 1000. Results are shown to be relatively stable when \textit{N} varies in this range, while become worst if bigger \textit{N} are used, as low informative features are also introduced. Santus et al. \shortcite{santus2016unsupervised} have also used LMI instead of PPMI as weighting function, but achieving lower results.\\
With respect to the limitations mentioned above for the Vector Cosine, APSyn has some advantages. First of all, it is by definition able to identify the extent of the intersection. Second, its sensibility to features with high values can be kept under control by tuning the value of \textit{N}. On top of it, feature values (i.e. their weights) do not affect directly the similarity score, as they are only used to build the feature rank. With reference to the toy-vectors presented above, APSyn would assign in fact completely different scores. The higher score would be assigned to \textit{$a$} and \textit{$d$}, as they share two relevant features out of three. The second higher score would be assigned to \textit{$a$} and \textit{$c$}, for the same reason as above. The lower score would be instead assigned to \textit{$a$} and \textit{$b$}, as they only share one non-salient feature. In section 3.4, we briefly discuss the hubness problem.

\subsection{Datasets}
For our evaluation, we used three widely popular datasets: WordSim-353 \cite{finkelstein2001placing}, MEN \cite{bruni2014multimodal}, SimLex-999 \cite{hill2015simlex}. These datasets have a different history, but all of them consist in word pairs with an associated score, that should either represent word association or word similarity.
WordSim-353 \cite{finkelstein2001placing} was proposed as a word similarity dataset containing 353 pairs annotated with scores between 0 and 10. However, Hill et al. \shortcite{hill2015simlex} claimed that the instructions to the annotators were ambiguous with respect to similarity and association, so that the subjects assigned high similarity scores to entities that are only related by virtue of frequent association (e.g. \textit{coffee} and \textit{cup}; \textit{movie} and \textit{theater}). On top of it, WordSim-353 does not provide the POS-tags for the 439 words that it contains, forcing the users to decide which POS to assign to the ambiguous words (e.g. [\textit{white}, \textit{rabbit}] and [\textit{run}, \textit{marathon}]). An extension of this dataset resulted from the subclassification carried out by Agirre et al. \shortcite{agirre2009study}, which discriminated between similar and associated word pairs. Such discrimination was done by asking annotators to classify all pairs according to the semantic relation they hold (i.e. identical, synonymy, antonymy, hypernymy, meronymy and none-of-the-above). The annotation was then used to group the pairs in three categories: similar pairs (those classified as identical, synonyms, antonyms and hypernyms), associated pairs (those classified as meronyms and none-of-the-above, with an average similarity greater than 5), and non-associated pairs (those classified as none-of-the-above, with an average similarity below or equal to 5). Two gold standard were finally produced: i) one for similarity, containing 203 word pairs resulting from the union of similar and non-associated pairs; ii) one for relatedness, containing 252 word pairs resulting from the union of associated and non-associated pairs. Even though such a classification made a clear distinction between the two types of relations (i.e. similarity and association), Hill et al. \shortcite{hill2015simlex} argue that these gold standards still carry the scores they had in WordSim-353, which are known to be ambiguous in this regard.\\
The MEN Test Collection \cite{bruni2014multimodal} includes 3,000 word pairs divided in two sets (one for training and one for testing) together with human judgments, obtained through Amazon Mechanical Turk. The construction was performed by asking subjects to rate which pair - among two of them - was the more related one (i.e. the most associated). Every pairs-couple was proposed only once, and a final score out of 50 was attributed to each pair, according to how many times it was rated as the most related. According to Hill et al. \shortcite{hill2015simlex}, the major weakness of this dataset is that it does not encode word similarity, but a more general notion of association.\\
SimLex-999 is the dataset introduced by Hill et al. \shortcite{hill2015simlex} to address the above mentioned criticisms of confusion between similarity and association. The dataset consists of 999 pairs containing 1,028 words, which were also evaluated in terms of POS-tags and concreteness. The pairs were annotated with a score between 0 and 10, and the instructions were strictly requiring the identification of word similarity, rather than word association. Hill et al. \shortcite{hill2015simlex} claim that differently from other datasets, SimLex-999 inter-annotator agreement has not been surpassed by any automatic approach.

\subsection{State of the Art Vector Space Models}

In order to compare our results with state-of-the-art DSMs, we report the scores for the Vector Cosines calculated on the neural language models (NLM) by Hill et al. \shortcite{hill2015simlex}, who used the code (or directly the embeddings) shared by the original authors. As we trained our models on almost the same corpora used by Hill and colleagues, the results are perfectly comparable.\\
The three models we compare our results to are: i) the convolutional neural network of Collobert and Weston \shortcite{collobert2008unified}, which was trained on 852 million words of Wikipedia; ii) the neural network of Huang et al. \shortcite{huang2012improving}, which was trained on 990 million words of Wikipedia; and iii) the word2vec of Mikolov et al. \shortcite{mikolov2013efficient}, which was trained on 1000 million words of Wikipedia and on the RCV Vol. 1 Corpus \cite{lewis2004rcv1}.

\begin{center}
\begin{table*}[ht]
\centering
\begin{tabular}{|c|c|c|c|c|c|c|}
\hline
\textbf{Dataset}         & \multicolumn{2}{c|}{\textbf{SimLex-999}} & \multicolumn{2}{c}{\textbf{WordSim-353}} & \multicolumn{2}{|c|}{\textbf{MEN}} \\
\hline
\textbf{Window}                                  & \textbf{2}                          & \textbf{3}                         & \textbf{2}                         & \textbf{3}                    & \textbf{2}                      & \textbf{3}                      \\
\hline
Cos Freq        & 0.149                      & 0.133                     & 0.172                      & 0.148                      & 0.089                  & 0.096                  \\
\hline
Cos LMI         & 0.248                      & 0.259                     & 0.321                      & 0.32                       & 0.336                  & 0.364                  \\
\hline
Cos PPMI         & 0.284                      & 0.267                     & 0.41                       & 0.407                      & 0.424                  & 0.433                  \\
\hline
Cos SVD-Freq300 & 0.128                      & 0.127                     & 0.169                      & 0.172                      & 0.076                  & 0.084                  \\
\hline
Cos SVD-LMI300  & 0.19                       & 0.21                      & 0.299                      & 0.29                       & 0.275                  & 0.286                  \\
\hline
\textbf{Cos SVD-PPMI300}  & \textbf{0.386}                     & \textbf{0.382}                     & \textbf{0.485}                      & \textbf{0.47}                       & \textbf{0.509}                  & \textbf{0.538}                  \\
\hline
APSynLMI-1000      & 0.18                       & 0.163                     & 0.254                      & 0.237                      & 0.205                  & 0.196                  \\
\hline
APSynLMI-500       & 0.199                      & 0.164                     & 0.283                      & 0.265                      & 0.226                  & 0.214                  \\
\hline
APSynLMI-100       & 0.206                      & 0.182                     & 0.304                      & 0.265                      & 0.23                   & 0.209                  \\
\hline
APSynPPMI-1000      & 0.254                      & 0.304                     & 0.399                      & 0.453                      & 0.369                  & 0.415                  \\
\hline
\textbf{APSynPPMI-500}       & 0.295                     & 0.32                      & \textbf{0.455}                      & \textbf{0.468}                      & 0.423                 & 0.478                \\
\hline
\textbf{APSynPPMI-100}       & \textbf{0.332}                      & \textbf{0.328}                     & 0.425                      & 0.422                      & \textbf{0.481}                  & \textbf{0.513}                  \\
\hline
\multicolumn{7}{|c|}{\textbf{State of the Art}}                                                                                                                                                 \\
\hline
            Mikolov et al.	    & \multicolumn{2}{c|}{0.282}                              & \multicolumn{2}{|c|}{0.442}                               & \multicolumn{2}{|c|}{0.433}     \\
\hline
\end{tabular}
\caption{Spearman correlation scores for our eight models trained on RCV Vol. 1, in the three datasets Simlex-999, WordSim-353 and MEN. In the bottom the performance of the state-of-the-art model of Mikolov et al. \shortcite{mikolov2013efficient}, as reported in Hill et al. \shortcite{hill2015simlex}.}
\label{my-label}
\end{table*}
\end{center}

\begin{center}
\begin{table*}[ht]
\centering
\begin{tabular}{|c|c|c|c|c|c|c|}
\hline
\textbf{Dataset}         & \multicolumn{2}{c|}{\textbf{SimLex-999}} & \multicolumn{2}{c}{\textbf{WordSim-353}} & \multicolumn{2}{|c|}{\textbf{MEN}} \\
\hline
\textbf{Window}                                  & \textbf{2}                          & \textbf{3}                         & \textbf{2}                         & \textbf{3}                    & \textbf{2}                      & \textbf{3}                      \\
\hline
Cos Freq                                & 0.148                      & 0.159                     & 0.199                     & 0.207                & 0.178                  & 0.197                  \\
\hline
Cos LMI                                 & 0.367                      & 0.374                     & 0.489                     & 0.529                & 0.59                   & 0.63                   \\
\hline
Cos PPMI                                 & 0.395                      & 0.364                     & 0.605                     & 0.622                & 0.733                  & 0.74                   \\
\hline
Cos SVD-Freq300                         & 0.157                      & 0.184                     & 0.159                     & 0.172                & 0.197                  & 0.226                  \\
\hline
Cos SVD-LMI300                          & 0.327                      & 0.329                     & 0.368                     & 0.408                & 0.524                  & 0.563                  \\
\hline
\textbf{Cos SVD-PPMI300}                          & \textbf{0.477}                      & \textbf{0.464}                     & \textbf{0.533}                    & \textbf{0.562}                & \textbf{0.769}                  & \textbf{0.779}                  \\
\hline
APSynLMI-1000                              & 0.343                      & 0.344                     & 0.449                     & 0.477                & 0.586                  & 0.597                  \\
\hline
APSynLMI-500                               & 0.339                      & 0.342                     & 0.438                     & 0.47                 & 0.58                   & 0.588                  \\
\hline
APSynLMI-100                               & 0.303                      & 0.31                      & 0.392                     & 0.428                & 0.48                   & 0.498                  \\
\hline
APSynPPMI-1000                              & 0.434                      & 0.419                     & 0.599                     & 0.643                & 0.749                  & 0.772                  \\
\hline
\textbf{APSynPPMI-500}         & \textbf{0.442}       & \textbf{0.423}       & \textbf{0.602}         & \textbf{0.653}     & \textbf{0.757}     & \textbf{0.773}     \\
\hline
APSynPPMI-100                               & 0.316                      & 0.281                     & 0.58                      & 0.608                & 0.703                  & 0.722                  \\
\hline
\multicolumn{7}{|c|}{\textbf{State of the Art}}                                                                                                                                                                  \\
\hline
Huang et al.          & \multicolumn{2}{c|}{0.098}                              & \multicolumn{2}{c}{0.3}                          & \multicolumn{2}{|c|}{0.433}                       \\
\hline
Collobert \& Weston & \multicolumn{2}{c|}{0.268}  & \multicolumn{2}{c}{0.494} & \multicolumn{2}{|c|}{0.575} \\
\hline
Mikolov et al.      & \multicolumn{2}{c|}{0.414}  & \multicolumn{2}{c}{0.655} & \multicolumn{2}{|c|}{0.699} \\
\hline
\end{tabular}
\caption{Spearman correlation scores for our eight models trained on Wikipedia, in the three datasets Simlex-999, WordSim-353 and MEN. In the bottom the performance of the state-of-the-art models of Collobert and Weston \shortcite{collobert2008unified}, Huang et al. \shortcite{huang2012improving}, Mikolov et al. \shortcite{mikolov2013efficient}, as reported in Hill et al. \shortcite{hill2015simlex}.}
\label{my-label}
\end{table*}
\end{center}

\section{Experiments}
In this section, we describe our experiments, starting from the training corpora (Section 3.1), to move to the implementation of twenty-eight DSMs (Section 3.2), following with the application and evaluation of the measures (Section 3.3), up to the performance analysis (Section 3.4) and the scalability test (Section 3.5).

\subsection{Corpora and Preprocessing}
We used two different corpora for our experiments: RCV vol. 1 \cite{lewis2004rcv1} and the Wikipedia corpus \cite{baroni2009wacky}, respectively containing 150 and 820 million words. The RCV Vol. 1 and Wikipedia were automatically tagged, respectively, with the POS tagger described in Dell'Orletta \shortcite{dell2009ensemble} and with the TreeTagger \cite{schmid1994probabilistic}.

\subsection{DSMs}
For our experiments, we implemented twenty-eight DSMs, but for reasons of space only sixteen of them are reported in the tables. All of them include the pos-tagged target words used in the three datasets (i.e. MEN, WordSim-353 and SimLex-999) and the pos-tagged contexts having frequency above 100 in the two corpora.
We considered as contexts the content words (i.e. nouns, verbs and adjectives) within a window of 2, 3 and 5, even though the latter was given up for its poor performances.\\
As for SVD factorization, we found out that the best results were always achieved when the number of latent dimensions was between 300 and 500. We report here only the scores for \textit{k} = 300, since 300 is one of the most common choices for the dimensionality of SVD-reduced spaces and it is always close to be an optimal value for the parameter.\\
Fourteen out of twenty-eight models were developed for RCV1, while the others were developed for Wikipedia. For each corpus, the models differed according to the window size (i.e. 2 and 3), to the statistical association measure used as a weighting scheme (i.e. none, PPMI and LMI) and to the application of SVD to the previous combinations.\\ 

\begin{center}
\begin{table*}[ht]
\centering
\begin{tabular}{|c|c|c|c|c|}
\hline
\textbf{Dataset} & \multicolumn{2}{c|}{\textbf{WSim (SIM)}} & \multicolumn{2}{c|}{\textbf{WSim (REL)}} \\ \hline
\textbf{Window}                                  & \textbf{2}                          & \textbf{3}                         & \textbf{2}                         & \textbf{3}      \\ \hline
Cos Freq & 0.208 & 0.158 & 0.167 & 0.175 \\ \hline
Cos LMI & 0.416 & 0.395 & 0.251 & 0.269 \\ \hline
Cos PPMI & 0.52 & 0.496 & 0.378 & 0.396 \\ \hline
Cos SVD-Freq300 & 0.240 & 0.214 & 0.051 & 0.084 \\ \hline
Cos SVD-LMI300 & 0.418 & 0.393 & 0.141 & 0.151 \\ \hline
\textbf{Cos SVD-PPMI300} & \textbf{0.550} & \textbf{0.522} & \textbf{0.325} & \textbf{0.323} \\ \hline
APSynLMI-1000 & 0.32 & 0.29 & 0.259 & 0.241 \\ \hline
APSynLMI-500 & 0.355 & 0.319 & 0.261 & 0.284 \\ \hline
APSynLMI-100 & 0.388 & 0.335 & 0.233 & 0.27 \\ \hline
\textbf{APSynPPMI-1000} & 0.519 & 0.525 & 0.337 & \textbf{0.397} \\ \hline
\textbf{APSynPPMI-500} & \textbf{0.564} & \textbf{0.546} & \textbf{0.361} & 0.382 \\ \hline
PMI APSynPPMI-100 & 0.562 & 0.553 & 0.287 & 0.309 \\ \hline
\end{tabular}
\caption{Spearman correlation scores for our eight models trained on RCV1, in the two subsets of WordSim-353.}
\label{my-label}
\end{table*}
\end{center}

\begin{center}
\begin{table*}[ht]
\centering
\begin{tabular}{|c|c|c|c|c|}
\hline
\textbf{Dataset} & \multicolumn{2}{c|}{\textbf{WSim (SIM)}} & \multicolumn{2}{c|}{\textbf{WSim (REL)}} \\ \hline
\textbf{Window}                                  & \textbf{2}                          & \textbf{3}                         & \textbf{2}                         & \textbf{3}      \\ \hline
Cos Freq & 0.335 & 0.334 & 0.03 & 0.05 \\ \hline
Cos LMI & 0.638 & 0.663 & 0.293 & 0.34 \\ \hline
Cos PPMI & 0.672 & 0.675 & 0.441 & 0.446 \\ \hline
Cos SVD-Freq300 & 0.35 & 0.363 & -0.013 & 0.001 \\ \hline
Cos SVD-LMI300 & 0.604 & 0.626 & 0.222 & 0.286 \\ \hline
\textbf{Cos SVD-PPMI300} & \textbf{0.72} & \textbf{0.725} & \textbf{0.444} & \textbf{0.486} \\ \hline
APSynLMI-1000 & 0.609 & 0.609 & 0.317 & 0.36 \\ \hline
APSynLMI-500 & 0.599 & 0.601 & 0.289 & 0.344 \\ \hline
APSynLMI-100 & 0.566 & 0.574 & 0.215 & 0.271 \\ \hline
APSynPPMI-1000 & 0.692 & 0.726 & 0.507 & 0.568 \\ \hline
\textbf{APSynPPMI-500} & \textbf{0.699} & \textbf{0.742} & \textbf{0.508} & \textbf{0.571} \\ \hline
APSynPPMI-100 & 0.66 & 0.692 & 0.482 & 0.516 \\ \hline
\end{tabular}
\caption{Spearman correlation results for our eight models trained on Wikipedia, in the subsets of WordSim-353.}
\label{my-label}
\end{table*}
\end{center}

\subsection{Measuring Word Similarity and Relatedness}
Given the twenty-eight DSMs, for each dataset we have measured the Vector Cosine and APSyn between the words in the test pairs. \\ The Spearman correlation between our scores and the gold standard was then computed for every model and it is reported in Table 1 and Table 2. In particular, Table 1 describes the performances on SimLex-999, WordSim-353 and MEN for the measures applied on RCV Vol. 1 models. Table 2, instead, describes the performances of the measures on the three datasets for the Wikipedia models. Concurrently, Table 3 and Table 4 describe the performances of the measures respectively on the RCV Vol. 1 and Wikipedia models, tested on the subsets of WordSim-353 extracted by Agirre et al. \shortcite{agirre2009study}.

\subsection{Performance Analysis}
Table 1 shows the Spearman correlation scores for Vector Cosine and APSyn on the three datasets for the eight most representative DSMs built using RCV Vol. 1. Table 2 does the same for the DSMs built using Wikipedia. For the sake of comparison, we also report the results of the state-of-the-art DSMs mentioned in Hill et al. \shortcite{hill2015simlex} (see Section 2.5).\\
With a glance at the tables, it can be easily noticed that the measures perform particularly well in two models: i) APSyn, when applied on the PPMI-weighted DSM (henceforth, APSynPPMI); ii) Vector Cosine, when applied on the SVD-reduced PPMI-weighted matrix (henceforth, CosSVDPPMI). These two models perform consistently and in a comparable way across the datasets, generally outperforming the state-of-the-art DSMs, with an exception for the Wikipedia-trained models in WordSim-353.\\
Some further observations are: i) corpus size strongly affects the results; ii) PPMI strongly outperforms LMI for both Vector Cosine and APSyn; iii) SVD boosts the Vector Cosine, especially when it is combined with PPMI; iv) \textit{N} has some impact on the performance of APSyn, which generally achieves the best results for \textit{N}=500. As a note about iii), the results of using SVD jointly with LMI spaces are less predictable than when combining it with PPMI.\\
Also, we can notice that the smaller window (i.e. 2) does not always perform better than the larger one (i.e. 3). The former appears to perform better on SimLex-999, while the latter seems to have some advantages on the other datasets. This might depend on the different type of similarity encoded in SimLex-999 (i.e. genuine similarity). On top of it, despite Hill et al. \shortcite{hill2015simlex}'s claim that no evidence supports the hypothesis that smaller context windows improve the ability of models to capture similarity \cite{agirre2009study,kiela2014systematic}, we need to mention that window 5 was abandoned because of its low performance.\\
With reference to the hubness effect, we have conducted a pilot study inspired to the one carried out by Schnabel et al. \shortcite{schnabel2015evaluation}, using the words of the SimLex-999 dataset as query words and collecting for each of them the top 1000 nearest neighbors. Given all the neighbors at rank \textit{r}, we have checked their rank in the frequency list extracted from our corpora. Figure 1 shows the relation between the rank in the nearest neighbor list and the rank in the frequency list. It can be easily noticed that the highest ranked nearest neighbors tend to have higher rank also in the frequency list, supporting the idea that frequent words are more likely to be nearest neighbors. APSyn does not seem to be able to overcome such bias, which seems to be in fact an inherent property of the DSMs \cite{radovanovic2010existence}. Further investigation is needed to see whether variations of APSyn can tackle this problem.\\

\begin{center}
  \includegraphics[height=40mm,width=80mm]{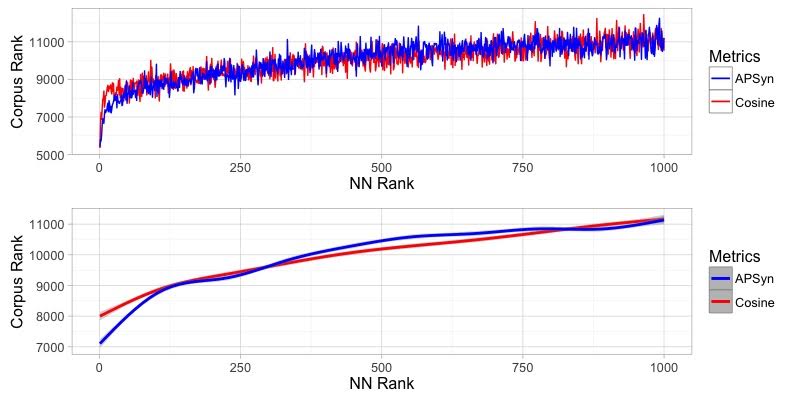}
  \captionof{figure}{Rank in the corpus-derived frequency list for the top 1000 nearest neighbors of the terms in SimLex-999, computed with Cosine (red) and APSyn (blue). The smoothing chart in the bottom uses the Generalized Additive Model (GAM) from the \textit{mgcv} package in \textit{R}.}
\end{center}

Finally, few words need to be spent with regard to the ability of calculating genuine similarity, as distinguished from word relatedness \cite{turney2001mining,agirre2009study,hill2015simlex}. Table 3 and Table 4 show the Spearman correlation scores for the two measures calculated on the models respectively trained on RCV1 and Wikipedia, tested on the subsets of WordSim-353 extracted by Agirre et al. \shortcite{agirre2009study}. It can be easily noticed that our best models work better on the similarity subset. In particular, APSynPPMI performs about 20-30\% better for the similarity subset than for the relatedness one (see Table 3), as well as both APSynPPMI and CosSVDPPMI do in Wikipedia (see Table 4).

\subsection{Scalability}
In order to evaluate the scalability of APSyn, we have performed a pilot test on WordSim-353 and MEN with the same corpus used by Baroni et al. \shortcite{baroni2014don}, which consists of about 2.8B words (i.e. about 3 times Wikipedia and almost 20 times RCV1). The best scores were obtained with APSyn, \textit{N}=1000, on a 2-window PPMI-weighted DSM. In such setting, we obtain a Spearman correlation of 0.72 on WordSim and 0.77 on MEN. These results are much higher than those reported by Baroni et al. \shortcite{baroni2014don} for the count-based models (i.e. 0.62 on WordSim and 0.72 on MEN) and slightly lower than those reported for the predicting ones (i.e. 0.75 on WordSim and 0.80 on MEN).

\section{Conclusions}
In this paper, we have presented the first systematic evaluation of APSyn, comparing it to Vector Cosine in the task of word similarity identification. We developed twenty-eight count-based DSMs, each of which implementing different hyperparameters. PPMI emerged as the most efficient association measure: it works particularly well with Vector Cosine, when combined with SVD, and it boosts APSyn.\\
APSyn showed extremely promising results, despite its conceptual simplicity. It outperforms the Vector Cosine in almost all settings, except when the latter is used on a PPMI-weighed SVD-reduced DSM. Even in this case, anyway, its performance is very competitive. Interestingly, our best models achieve results that are comparable to - or even better than - those reported by Hill et al. \shortcite{hill2015simlex} for the state-of-the-art word embeddings models. In Section 3.5 we show that APSyn is scalable, outperforming the state-of-the-art count-based models reported in Baroni et al. \shortcite{baroni2014don}.\
On top of it, APSyn does not suffer from some of the problems reported for the Vector Cosine, such as the inability of identifying the number of shared features. It still however seems to be affected by the hubness issue, and more research should be carried out to tackle it.\
Concerning the discrimination between similarity and association, the good performance of APSyn on SimLex-999 (which was built with a specific attention to genuine similarity) and the large difference in performance between the two subsets of WordSim-353 described in Table 3 and Table 4 make us conclude that APSyn is indeed efficient in quantifying genuine similarity.\\
To conclude, being a linguistically and cognitively grounded metric, APSyn offers the possibility for further improvements, by simply combining it to other properties that were not yet considered in its definition. A natural extension would be to verify whether APSyn hypothesis and implementation holds on SVD reduced matrices and word embeddings.

\section*{Acknowledgments}

This paper is partially supported by HK PhD Fellowship Scheme, under PF12-13656. Emmanuele Chersoni's research is funded by a grant of the University Foundation A*MIDEX. Thanks to Davis Ozols for the support with \textit{R}.

\end{document}